\documentclass[letterpaper, 10 pt, journal, twoside]{IEEEtran}
\usepackage{times}
\usepackage[pdftex]{graphicx}
\usepackage{amsmath,amssymb,amsopn,amstext,amsfonts}
\usepackage{cancel}
\usepackage[space]{cite}
\usepackage{pdfsync}
\usepackage{balance}
\usepackage{color}
\usepackage{mathtools}
\usepackage{algpseudocode}
\usepackage{algorithm} %\usepackage[ruled,vlined,linesnumbered]{algorithm2e}
\usepackage{bm}
\usepackage{bbm}

\usepackage{float}
\usepackage{epstopdf}
\usepackage{pifont}
\usepackage{fixltx2e}
\usepackage{amsmath}
\usepackage{multirow}
\usepackage{url}
\usepackage{verbatim}
\usepackage{siunitx} 
\usepackage{epstopdf}

\usepackage{bbm}
\usepackage{bm}

\usepackage{diagbox}
\usepackage{float}
\usepackage{epstopdf}
\usepackage{pifont}
\usepackage{fixltx2e}
\usepackage{multirow}
\usepackage{url}
\usepackage{verbatim}
\usepackage{siunitx} 
\usepackage{epstopdf}
\usepackage{bbding}

\usepackage{times}
\usepackage{cancel}
\usepackage{pdfsync}
\usepackage{balance}
\usepackage{color}
\usepackage{mathtools}

\usepackage[linkcolor=black,citecolor=black,urlcolor=black,colorlinks=true]{hyperref}
\usepackage{booktabs}
\usepackage{makecell}
\usepackage{enumerate}
\usepackage{arydshln}
\usepackage[dvipsnames]{xcolor}
\usepackage{anyfontsize}

\usepackage{xr}
\usepackage{cleveref}
\definecolor{darkred}{RGB}{195,25,25}

\title{H3-Mapping: Quasi-Heterogeneous Feature Grids for Real-time Dense Mapping Using Hierarchical Hybrid Representation}

\author{Chenxing Jiang$^{2}$, Yiming Luo$^{1}$, Boyu Zhou$^{1,\dag}$, Shaojie Shen$^{2}$
	\vspace{-0.5cm}
	\thanks{Manuscript received: March, 15, 2024; Revised June, 3, 2024; Accepted July, 7, 2024. This paper was recommended for publication by Editor Javier Civera upon evaluation of the Associate Editor and Reviewers' comments.}
	\thanks{\textsuperscript{\dag} \textbf{Corresponding Author}. }
	\thanks{\textsuperscript{1} The School of Artificial Intelligence, Sun Yat-Sen University, Zhuhai.}
	\thanks{\textsuperscript{2} Department of Electronic and Computer Engineering, The Hong Kong 	University of Science and Technology, Hong Kong, China.}
	\thanks{\scriptsize \{\href{mailto:cjiangan@connect.ust.hk}{cjiangan@connect.ust.hk}, \href{mailto:zhouby23@mail.sysu.edu.cn}{zhouby23@mail.sysu.edu.cn}\}}
	\thanks{Digital Object Identifier (DOI): see top of this page.}
}
\begin{document}
	
	\maketitle

	\begin{abstract}   
		In recent years, implicit online dense mapping methods have achieved high-quality reconstruction results, showcasing great potential in robotics, AR/VR, and digital twins applications. However, existing methods struggle with slow texture modeling which limits their real-time performance.
		To address these limitations, we propose a NeRF-based dense mapping method that enables faster and higher-quality reconstruction. To improve texture modeling, we introduce quasi-heterogeneous feature grids, which inherit the fast querying ability of uniform feature grids while adapting to varying levels of texture complexity. Additionally, we present a gradient-aided coverage-maximizing strategy for keyframe selection that enables the selected keyframes to exhibit a closer focus on rich-textured regions and a broader scope for weak-textured areas.
		Experiments show that our method surpasses existing implicit approaches in texture fidelity, geometry accuracy, and time consumption. The code for our method will be available at: https://github.com/SYSU-STAR/H3-Mapping.	
	\end{abstract}
	\begin{IEEEkeywords}
		Mapping; RGB-D Perception; Visual Learning
	\end{IEEEkeywords}
	
	\IEEEpeerreviewmaketitle
	\vspace{-0.4cm}
	\section{Introduction}
\begin{figure}[t!]
		\centering
		\includegraphics[width=1\linewidth]{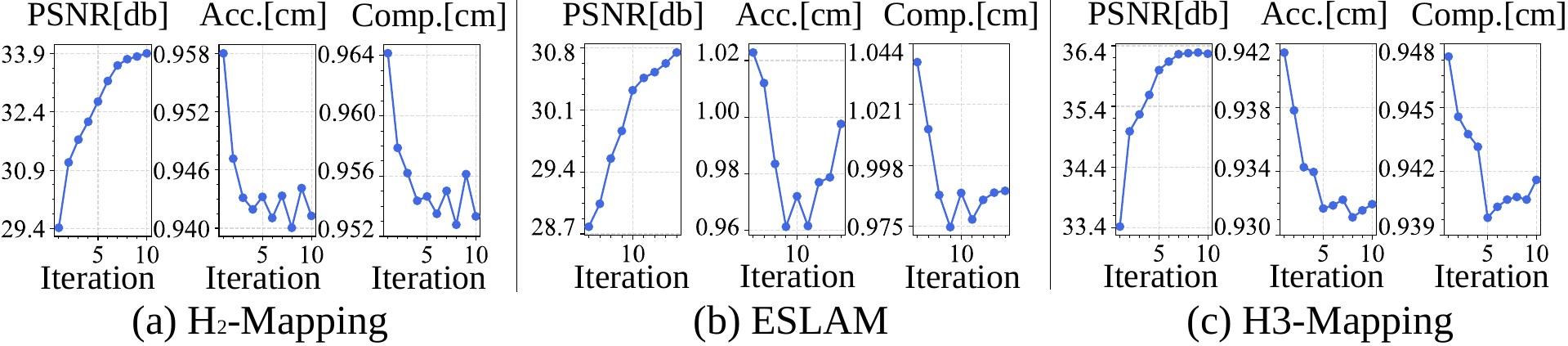}
		\caption{
			Mapping results of two recent NeRF-based dense mapping methods\cite{jiang2023h2,johari2022eslam}(a,b) and our proposed method(c) on Room1 in the Replica dataset\cite{straub2019replica} with varying training iterations per frame. It reveals that the geometry error (measured as accuracy(Acc.) and completion(Comp.)) reaches a low value with limited iterations. In methods\cite{jiang2023h2,johari2022eslam}, the texture (evaluated by PSNR) requires further training, but our method achieves higher PSNR with fewer iterations by rapidly converging. The metrics are defined in Sec.\ref{subsubsec:metrics}.}
		\label{fig:iter_curve}
		\vspace{-0.5cm}
	\end{figure}

	\vspace{-0.1cm}
	\IEEEPARstart{R}{eal-time} reconstruction of high-quality scene geometry and texture holds significant importance in robotics, AR/VR, and digital twins applications. 
	Following the advent of Neural Radiance Fields (NeRF)\cite{mildenhall2021nerf}, several recent works\cite{sucar2021imap,zhu2022nice,zhu2024nicer,yang2022vox,johari2022eslam,jiang2023h2,sandstrom2023point} have employed implicit scene representations to achieve compact and higher-fidelity mapping results compared to explicit scene representations like occupancy grids\cite{hornung2013octomap}, TSDF \cite{newcombe2011kinectfusion}, surfels\cite{wang2019real}, and meshes\cite{ruetz2019ovpc}. 
	However, NeRF-based dense mapping methods are hindered by long training times. 
	As shown in Fig. \ref{fig:iter_curve}, optimizing textures is slower than the geometry, making it a bottleneck in these methods. 
	
\begin{figure}[t!]
		\centering
		\includegraphics[width=1\linewidth]{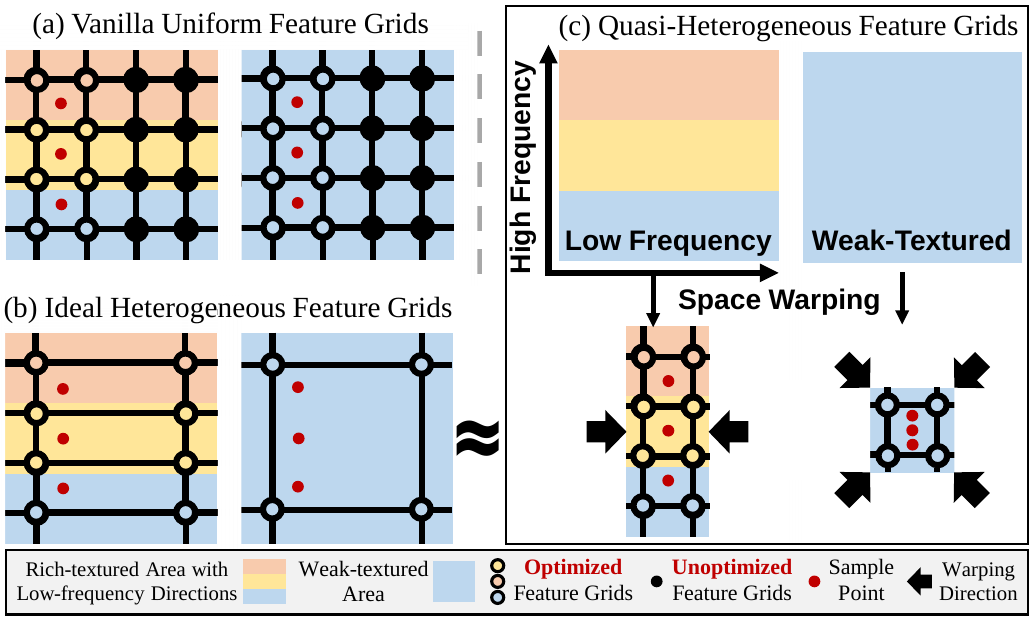}
		\caption{The illustration of quasi-heterogeneous feature grids.}
		\label{fig:Quasi_Hetero_Gird}
		\vspace{-1.6cm}
	\end{figure}

	Uniform feature grids have been used in various NeRF-based mapping methods\cite{zhu2022nice,zhu2024nicer, yang2022vox, johari2022eslam, jiang2023h2}, valued for their rapid querying properties. In these methods, tiny MLPs are used to decode the scene information stored in feature grids. 
	To model textures, these uniform grids treat the texture complexity in a spatially homogeneous manner. However, texture complexity is nonuniform across various scenes. For instance, walls commonly comprise regions with low texture complexity, while blinds, grout lines, and wooden furniture often exhibit stripes with both high-frequency directions (strong color variations) and low-frequency directions (weak color variations).
	Using the texture in Fig. \ref{fig:Quasi_Hetero_Gird}(c) as an illustration, the left-upper figure showcases a rich-textured area with a horizontal low-frequency direction, while the right-upper figure represents a weak-textured area. 
	Due to the shared MLP decoder and the spectral bias of MLPs\cite{rahaman2019spectral}, similar feature grids tend to model areas with similar colors.
	Therefore, in methods\cite{zhu2022nice,zhu2024nicer, yang2022vox,johari2022eslam, jiang2023h2} that employ uniform feature grids, allocating a large number of feature grids in the low-frequency direction and weak-textured areas can be redundant. 
	As optimization updates only propagate to grids where the samples are located, encoding the scene with more feature grids requires more training time. This makes it hard to sufficiently optimize all grids with limited sampling and training time in online mapping tasks.
	To better model textures with various levels of complexity in the scene, isotropic feature points\cite{sandstrom2023point} and anisotropic gaussians\cite{keetha2023splatam} are proposed for dense mapping. However, feature points\cite{sandstrom2023point} suffer from inefficient querying, and their isotropic nature hinders efficient modeling of textures with anisotropic complexity. Besides, while anisotropic gaussians generate realistic renderings, 
	they are unable to accurately represent the scene geometry. Additionally, the process of gaussian densification requires more training time, which reduces the real-time capability.
	
	Aiming to design a faster and more accurate mapping method, especially in terms of enhancing texture optimization efficiency, we propose quasi-heterogeneous feature grids, which inherit the fast querying ability of uniform grids while adapting to various levels of texture complexity. 
	As depicted in Fig. \ref{fig:Quasi_Hetero_Gird}(c), once the texture pattern is determined using the fast line detector\cite{suarez2022elsed} and color gradient, we employ space warping to compress the texture itself along the low-frequency directions for the rich-textured areas where there exist low-frequency patterns. On the other hand, for all of the weak-textured areas, we apply space warping to scale the texture itself to a smaller region. 
	This approach can be regarded as the approximation of the ideal heterogeneous feature grids as shown in Fig. \ref{fig:Quasi_Hetero_Gird}(b), which utilize irregular grids to adapt to different textures complexity but are inefficient to maintain. 
	Therefore, in the online mapping task where sampling and training time are limited, the entire region can be effectively trained. 
	we propose a gradient-aided coverage-maximizing strategy for keyframe selection, which adaptively addresses varying levels of texture complexity.
	This improvement enables selected keyframes to focus more closely on rich-textured areas and have a broader scope for weak-textured regions, allowing for including more complete scene information into the keyframe set. To maintain storage efficiency, an adaptive pruning technique is employed to remove redundant keyframes. 
	Additionally, for geometry reconstruction, we use uniform feature grids and incorporate the easy-to-optimize explicit octree SDF priors\cite{jiang2023h2} to create a hierarchical hybrid representation. 
	Feature grids for both texture and geometry are stored in hash maps with multiple resolutions. 
	In summary, the proposed work contributes as follows:

	\begin{itemize}
		\item The quasi-heterogeneous feature grids, which inherit the efficient querying capability of uniform grids while adapting to different levels of texture complexity, resulting in faster and more accurate texture modeling.
		
		\item The gradient-aided coverage-maximizing strategy for keyframe selection, which treats various texture complexities adaptively, leading to the full use of data from each chosen keyframe.
		
		\item Extensive experiments show our method can get superior mapping results with less runtime compared to existing implicit online mapping methods. 
	\end{itemize}
	
	\begin{figure*}[t!]
		\centering
		\includegraphics[width=1\textwidth]{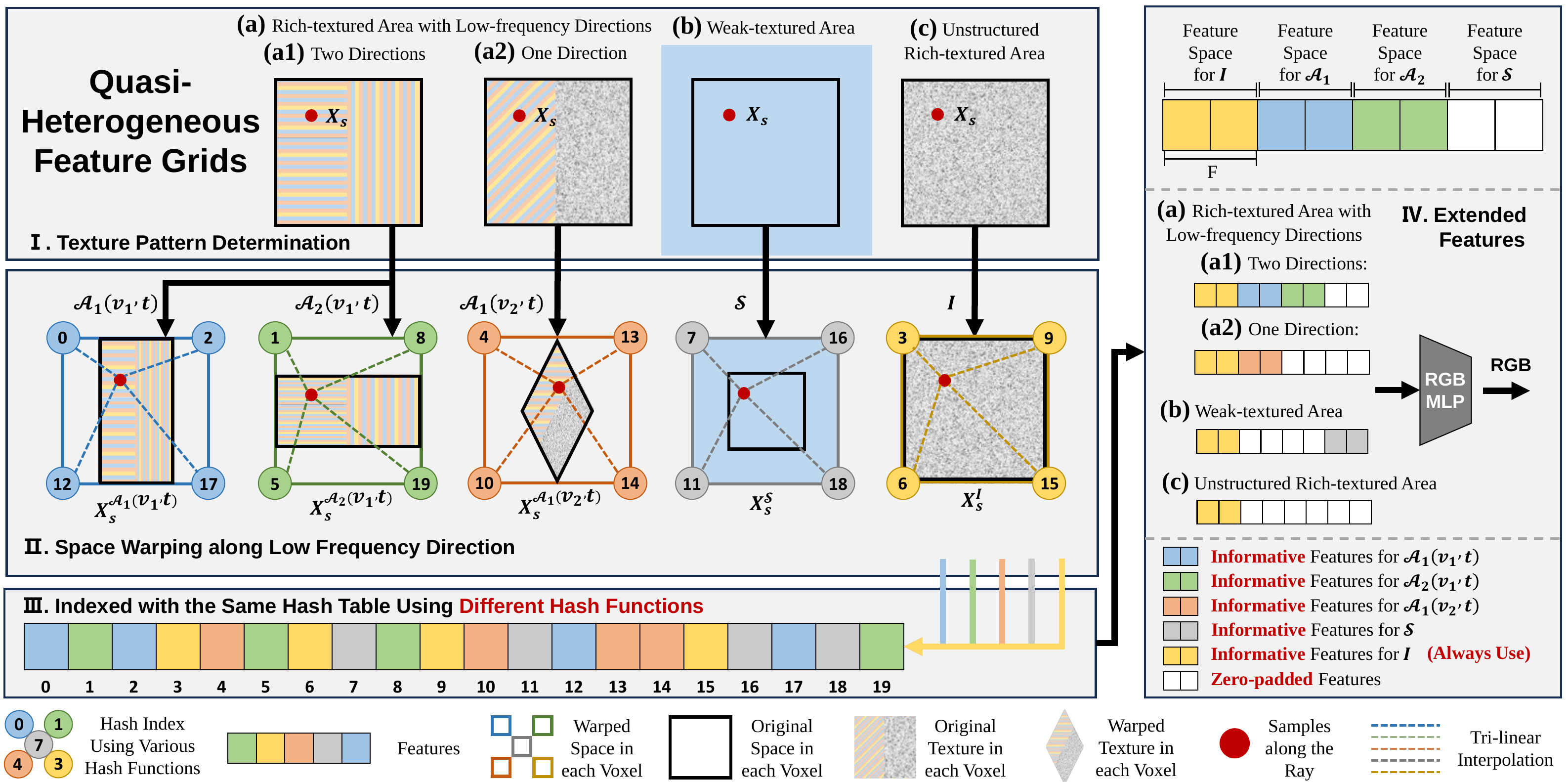}
		\caption{The illustration of the quasi-heterogeneous feature grids. For simplicity, we assume two maximum low-frequency directions ($M = 2$) and only show the features of one level in the multiresolution hash grids.}
		\label{fig:quasi_pipeline}
		\vspace{-0.7cm}
	\end{figure*}
	\vspace{-0.5cm}
	\section{Related Works}
	\subsection{Implicit Online Dense Mapping}
	Recently, several methods\cite{sucar2021imap, zhu2022nice, johari2022eslam, jiang2023h2, yang2022vox, sandstrom2023point} utilize latent features and neural networks as implicit representations for RGB-D dense mapping and yield more compact and accurate results. In the online implicit mapping pipeline, scene representation and keyframe selection strategy are two main components.
	\subsubsection{Scene Representation}
	iMap\cite{sucar2021imap} demonstrates, for the first time, that Multi-Layer Perceptrons (MLPs) with position encoding can serve as the sole scene representation in dense mapping tasks. 
	To enhance speed and expand representation capacity, various uniform grid representations have been introduced, including dense 3D grids\cite{zhu2022nice}, sparse octree grids\cite{yang2022vox}, factored grids\cite{johari2022eslam}, and multiresolution hash grids\cite{jiang2023h2}. These grid-based methods offer efficient local feature storage and enable fast neighborhood querying. As a result, they effectively reduce the size of MLPs, achieving state-of-the-art accuracy and training speed.
	However, the resolutions of feature grids are pre-defined and cannot be adjusted during mapping.
	Leveraging the inherent flexibility of point-based representations, Point-SLAM\cite{sandstrom2023point} employs a dynamic resolution approach guided by color gradients to allocate more points in complex regions. However, these feature points exhibit inefficient querying compared to feature grids, resulting in slower performance in online dense mapping tasks. Moreover, the isotropic nature hampers efficient modeling of textures with anisotropic complexity, as it often leads to an excessive allocation of points in the low-frequency direction of rich-textured areas. 
	On the other hand, SplaTAM\cite{keetha2023splatam} adopts anisotropic gaussians and employs the gaussian splatting technique\cite{kerbl20233d} to avoid the slow neighborhood querying in Point-SLAM\cite{sandstrom2023point}. However, the reconstructed geometry lacks the desired accuracy, and its gaussian densification process limits real-time capability.
	In contrast, our proposed quasi-heterogeneous feature grids combine the efficiency of grid-based representations with the flexibility of point-based representations. It exhibits adaptive behavior by using fewer feature grids in weak-textured areas and along the low-frequency direction in rich-textured regions. As a result, our method exhibits reduced training time and improved accuracy compared to previous approaches.
	\subsubsection{Keyframe Selection Strategy}
	In online implicit mapping, selecting keyframes to be retrained is crucial to mitigate the forgetting issue. iMap\cite{sucar2021imap} selects keyframes based on the distribution of loss values. However, it is time-consuming due to the need to recalculate rendered pixels. 
	In contrast, NICE-SLAM\cite{zhu2022nice} chooses keyframes that overlap with the current frame, reducing the number of trained parameters and maintaining static geometry outside the current view. However, it may not perform well in marginal areas where seldom observed. 
	H$_2$-Mapping\cite{jiang2023h2} uses the coverage-maximizing strategy to ensure all allocated voxels are covered with minimal iteration. However, merely selecting frames with maximum observed voxels may lead to viewing the rich-textured area from a large distance, which degrades mapping accuracy. To address this issue, we enhance the coverage-maximizing strategy\cite{jiang2023h2} with the aid of the color gradient. It enables the selected keyframes to have a closer focus on rich-textured areas and a broader scope for weak-textured regions.
	\vspace{-0.45cm}
	\subsection{Frequency-aware Feature Grids}
	
	To reduce aliasing caused by high-resolution feature grids, Zip-NeRF\cite{barron2023zip} proposes a down-weighting strategy to suppress unreliable high-frequency features.
	EvaSurf\cite{gao2023evasurf}, on the other hand, adopts a progressive grid approach with increasing bandwidth to reduce noise.
	Neuralangelo\cite{li2023neuralangelo} improves surface consistency by extending the optimization process beyond the local hash grid cell using numerical gradients.
	The aforementioned methods primarily address noise reduction in weak-textured regions caused by high-resolution grids by isotropic scaling the grid resolution. However, they do not tackle the training efficiency in detailed areas.
	In contrast, Zhan\cite{zhan2023general} introduces a learnable gauge network that transforms coordinates or hash indices to new ones that are better suited for scene modeling. Additionally, Yi\cite{yi2023canonical} proposes a set of canonicalizing transformations to reduce the axis-aligned bias of factored grid representations.
	While the learnable transformation can outperform pre-defined ones in offline scenarios, their additional learning time may not be efficient for online dense mapping tasks with limited sampling and training iterations. However, our quasi-heterogeneous feature grids, which incorporate space warping, allow for efficient transformation to adapt to various levels of texture complexity and enable rapid texture learning. Moreover, our approach differs from Mip-NeRF360\cite{barron2022mip}, MERF\cite{reiser2023merf}, and F$^{2}$-NeRF\cite{wang2023f2}, which only considers view distance during space warping and does not take the texture pattern into account.
	\vspace{-0.2cm}
	\section{Quasi-heterogeneous Feature Grids}
	\label{sec:quasi-h}
	When receiving a new frame, we allocate new voxels based on the provided pose and depth image, and then incrementally build a sparse voxel octree with a large voxel size that covers all visible areas.
	As shown in Fig. \ref{fig:quasi_pipeline}(\uppercase\expandafter{\romannumeral1}), we first classify the texture within each leaf node voxel of octree into three categories, following the subsequent priority order: (1) Rich-textured area with low-frequency directions; (2) Weak-textured area; and (3) Unstructured rich-textured area. For instance, the texture pattern in Fig. \ref{fig:quasi_pipeline}(a2) is considered as a rich-textured area with one low-frequency direction. Based on the texture pattern, we apply the corresponding space warping to construct quasi-heterogeneous feature grids as shown in Fig. \ref{fig:quasi_pipeline}(\uppercase\expandafter{\romannumeral2}). Then, we use different hash functions to alleviate the conflict problem, as described in Fig. \ref{fig:quasi_pipeline}(\uppercase\expandafter{\romannumeral3}). Lastly, as illustrated in Fig. \ref{fig:quasi_pipeline}(\uppercase\expandafter{\romannumeral4}), features from various warping functions are concatenated to be equal length and fed into a small MLP to predict the color. 
	\vspace{-0.7cm}
	\subsection{Texture Pattern Determination and Space Warping}
	\subsubsection{Rich-textured Area with Low-frequency Directions}
\begin{figure}[t!]
		\centering
		\includegraphics[width=1\linewidth]{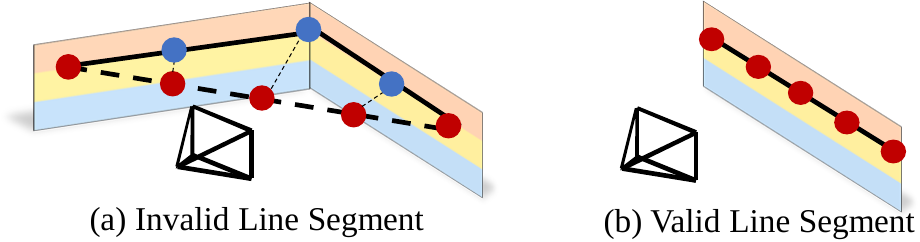}
		\caption{Illustration of the validation process for a line segment. Red points: Uniformly sampled points along the line connecting the two detected endpoints. Blue points: Back-projected points in the direction of the red points.}
		\label{fig:valid_line}
		\vspace{-0.7cm}
	\end{figure}

	When rich-textured regions exhibit a low-frequency direction, it is common for these areas to exhibit line segments. Therefore, we use the ELSED algorithm\cite{suarez2022elsed} to detect the line segments in the color image and then discard segments resulting from occlusions between objects based on the depth image. Additionally, as shown in Fig. \ref{fig:valid_line}, to ensure that the line segments lie on the same plane, we uniformly sample five points (red) along the segments in 3D space and project them onto the depth image, generating back-projected points (blue). Only if the sum of distances between the uniformly sampled points and the back-projected points is below a threshold $tr_{plane}$, is the line segment considered valid. 
	
\begin{figure}[t!]
		\centering
		\includegraphics[width=1\linewidth]{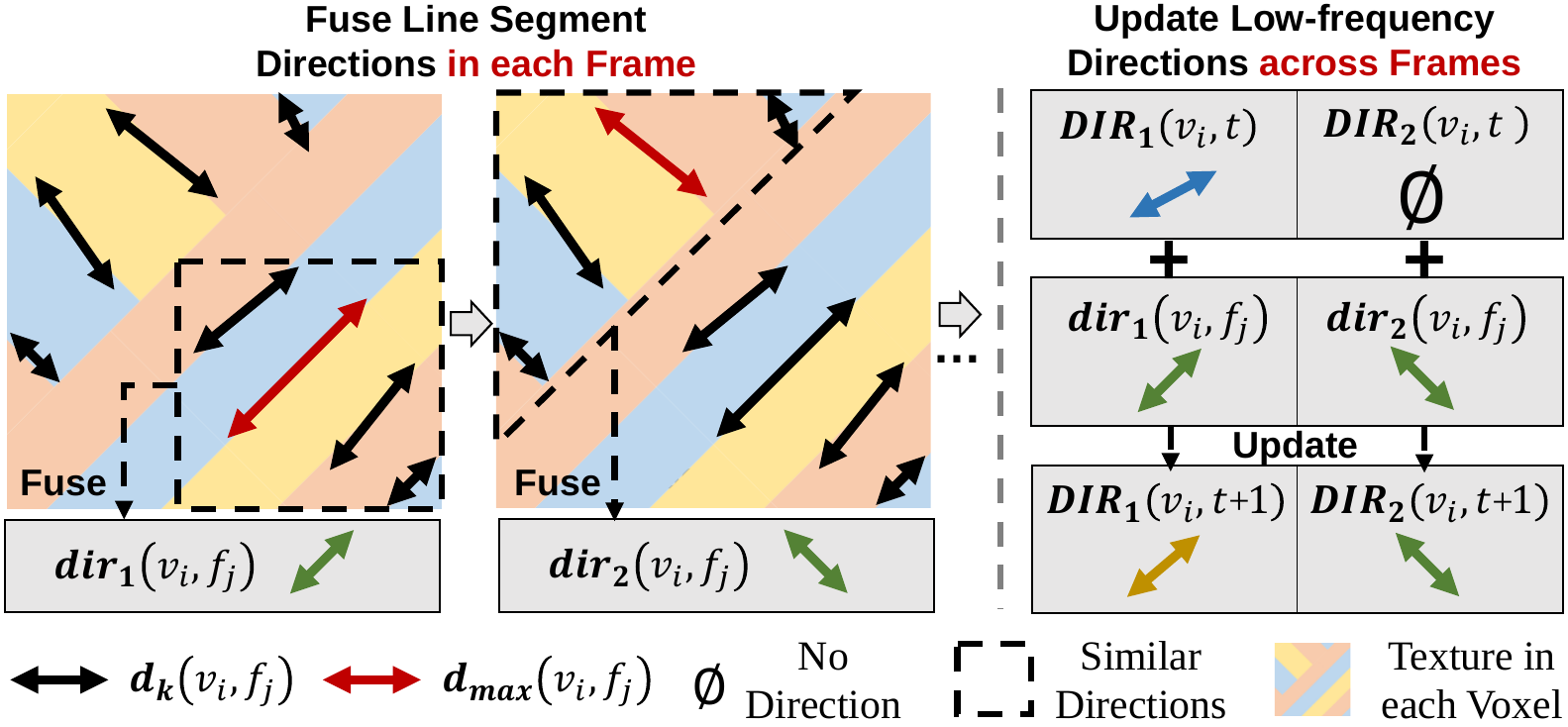}
		\caption{To find the low-frequency directions, we fuse the line directions of line segments that are similar to the one with the highest score in each frame and combine the results from multiple frames to obtain the updated directions.}
		\label{fig:line_fuse}
		\vspace{-1.5cm}
	\end{figure}

	Subsequently, we assume a maximum of $M$ low-frequency directions coexist within one voxel. As depicted in Fig. \ref{fig:line_fuse}, we fuse the 3D line directions within each voxel for each frame and use these results to update the low-frequency directions in each voxel across frames. Specifically, as depicted in the left figure of Fig. \ref{fig:line_fuse}, we select and fuse 3D line direction for each leaf node voxel $v_i$ of octree that intersects with line segments in the frame $f_j$. This process yields the fused 3D line direction $\mathbf{dir}_{n}(v_i, f_j)$ and its weight value $\mathbf{w}_{n}(v_i, f_j)$ as follows:
	\vspace{-0.1cm}
	\begin{equation}
		\footnotesize
		\resizebox{0.9\hsize}{!}{$
			\begin{aligned}
				\mathbf{dir}_{n}(v_i, f_j) &= \frac{\sum_{k \in \mathbf{U}} w_{k}(v_i, f_j) \mathbf{d}_{k}(v_i, f_j) }{\sum_{k \in \mathbf{U}} w_{k}(v_i, f_j)}, \\
				\mathbf{w}_{n}(v_i, f_j) &= \sum_{k \in \mathbf{U}} w_{k}(v_i, f_j), \ 
				\mathbf{U} = \{k \ | \  \langle \mathbf{d}_{k}(v_i, f_j) ,\mathbf{d}_{max}(v_i, f_j) \rangle \leq tr_{near}\} ,
			\end{aligned}$}
		\vspace{-0.1cm}
	\end{equation}
	where $\mathbf{d}_{k}(v_i, f_j)$ is the detected 3D line direction. The confidence scores from ELSED\cite{suarez2022elsed} are denoted as $w_{k}(v_i, f_j)$. Besides, $\mathbf{d}_{max}(v_i, f_j)$ is the 3D line direction with the highest score, and $tr_{near}$ is the threshold to identify line segments whose direction closely aligns with $\mathbf{d}_{max}(v_i, f_j)$. Afterwards, we remove all the $\mathbf{d}_{k \in \mathbf{U}}(v_i, f_j)$ from line direction set $\mathbf{U}$ and repeat the process until either there is no line direction belongs to $v_{i}$ or the selected direction number $n$ reaches $M$ in one voxel.
	During the mapping process, we keep track of $\mathbf{DIR}_{m}(v_i,t)$, which represents the $m^{th}$ low-frequency direction of each voxel $v_i$ at timestamp $t$. We update this direction using the fused line directions $\mathbf{dir}_{n}(v_i, f_j)$ from each frame $f_j$ as illustrated in the right figure of Fig. \ref{fig:line_fuse}. Specifically, 
	\begin{equation}
		\footnotesize
		\resizebox{0.9\hsize}{!}{$
			\begin{aligned}
				\gamma &= \mathbb{I}(\langle \mathbf{dir}_{n}(v_i, f_j) ,\mathbf{DIR}_{m}(v_i,t) \rangle \leq tr_{near}),\\
				\mathbf{DIR}_{m}(v_i,t\!+\!1) \!&=\! \frac{\mathbf{W}_{m}(v_i,t)  \mathbf{DIR}_{m}(v_i,t) \!+\! \gamma \mathbf{w}_{n}(v_i, f_j)  \mathbf{dir}_{n}(v_i, f_j) }{\mathbf{W}_{m}(v_i,t) \!+\!  \gamma \mathbf{w}_{n}(v_i, f_j)},\\
				\mathbf{W}_{m}(v_i,t\!+\!1) \!&=\! \mathbf{W}_{m}(v_i,t) \!+\! \gamma \mathbf{w}_{n}(v_i, f_j),
			\end{aligned}
			$}
		\vspace{-0.2cm}
	\end{equation}
	where $\mathbb{I}(\cdot)$ is the indicator function and $\mathbf{W}_{m}(v_i,t)$ is the weight value of $\mathbf{DIR}_{m}(v_i,t)$. If $m < M$, we include the $\mathbf{dir}_{n}(v_i, f_j)$ that is not similar to any $\mathbf{DIR}_{m}(v_i,t)$ to become the new low-frequency direction of voxel $v_i$.
	
	Once the low-frequency directions $\mathbf{DIR}_{m}(v_i,t)$ are determined, we apply the affine warping function $\mathcal{A}_{m}(v_i,t): \ \mathbb{R}^{3} \to \mathbb{R}^{3}$ to warp the space within each leaf node voxel around the voxel's center, as shown in Fig. \ref{fig:quasi_pipeline}(a). Specifically, for a given sample point $\mathbf{x}_p$ within the voxel $v_i$ and its center position denoted as $\mathbf{x}_c(v_i)$, the warped coordinate $\mathbf{x}_{p}^{\mathcal{A}_{m}(v_i,t)}$ corresponding to a specific low-frequency direction can be calculated as follows:
	\vspace{-0.5cm}
	\begin{equation}
		\footnotesize
		\resizebox{0.9\hsize}{!}{$
			\begin{aligned}
				&\mathbf{DIR}\! = \![b_{x}, b_{y}, b_{z}]^{\top} \!\in\! S^2, \ 
				\mathit{R}(\mathbf{DIR})\! = \!
				\begin{bmatrix}
					bz\! + \!\frac{{b_{y}^2}}{{1 + b_{z}}} & \frac{{-b_{x}b_{y}}}{{1 + b_{z}}} & b_{x} \\
					\frac{{-b_{x}b_{y}}}{{1 + b_{z}}} & 1\! - \! \frac{{b_{y}^2}}{{1 + b_{z}}} & b_{y} \\
					-b_{x} & -b_{y} & b_{z}
				\end{bmatrix}, \\
			\end{aligned}
			\vspace{-0.3cm}
			$}
	\end{equation}
	\vspace{-0.35cm}
	\begin{equation}
		\footnotesize
		\resizebox{0.9\hsize}{!}{$
			\begin{aligned}
				&\mathbf{A}_{m}(v_i,t) = 
				\begin{bmatrix}
					\begin{smallmatrix}
						1 & 0 & 0 \\
						0 & 1 & 0 \\
						0 & 0 & \mathcal{C}
					\end{smallmatrix}
				\end{bmatrix}		 \mathit{R}(\mathbf{DIR}_{m}(v_i,t))^{\top}, \\
				&\mathcal{A}_{m}(v_i,t): \ \mathbf{x}_{p}^{\mathcal{A}_{m}(v_i,t)} = \mathbf{A}_{m}(v_i,t)(\mathbf{x}_p - \mathbf{x}_c(v_i)) + \mathbf{x}_c(v_i),
			\end{aligned}
			$}
		\vspace{-0.1cm}
	\end{equation}
	where the 3D direction $\mathbf{DIR}$ is represented as a normalized vector $[b_{x}, b_{y}, b_{z}]^{\top}$, corresponding to a point on the two-dimensional sphere $S^2$. The function $\mathit{R}(\cdot)$ is the rotation matrix corresponding to the tilt of the 3D direction, and $\mathcal{C}<1$ is the compression rate. Intuitively, the space warping function can be considered as a two-step process. First, the space is rotated to align the low-frequency direction with the z-axis of the reference coordinate system. Then, compression is applied along the z-axis around $\mathbf{x}_c(v_i)$.
	If there are multiple low-frequency directions within the voxel, as depicted in Fig. \ref{fig:quasi_pipeline}(a1), we repeat this process for each direction and obtain multiple warped coordinates. 
	To ensure real-time performance, in this work, we employ linear space warping for stripe-like texture due to its efficient processing. However, it can be extended to recognize a broader range of texture patterns and apply corresponding non-linear warping functions.
	\subsubsection{Weak-textured Area}
	\label{subsubsec:weak}
	Excluding the previously determined rich-textured area with low-frequency directions, we identify the weak-textured area in each leaf node voxel $v_i$ based on the color gradient.
	To mitigate inaccurate determination of weak-textured areas due to limited field of view and occlusions, we fuse the color gradient in each voxel across multiple frames as follows:
	\begin{equation}
		\footnotesize
		\begin{aligned}
			&\mathbf{G}(v_i,t+1) = \frac{\mathbf{CNT}(v_i,t)  \mathbf{G}(v_i,t) + cnt(v_i, f_j)  g(v_i, f_j) }{\mathbf{CNT}(v_i,t) +  cnt(v_i, f_j)},\\
			&\mathbf{CNT}(v_i,t+1) = \mathbf{CNT}(v_i,t) + cnt(v_i, f_j),
		\end{aligned}
	\end{equation}
	where $g(v_i, f_j)$ is the average color gradient of all pixels within voxel $v_i$ in frame $f_j$ and $cnt(v_i, f_j)$ denotes the count of all pixels within voxel $v_i$ in frame $f_j$. Besides, $\mathbf{G}(v_i,t)$ and $\mathbf{CNT}(v_i,t)$ correspond to the tracked gradient value and pixel count of voxel $v_i$ at timestamp $t$. If the tracked gradient value, $\mathbf{G}(v_i,t)$ is lower than the threshold $tr_{weak}$, the region within voxel $v_i$ will be classified as the weak-textured area.
	
	Once the weak-textured areas are determined, we apply the scaling warping function $\mathcal{S}: \ \mathbb{R}^{3} \to \mathbb{R}^{3}$ to scale all the weak-textured areas altogether, as shown in Fig. \ref{fig:quasi_pipeline}(b). Specifically, for a given sample point $\mathbf{x}_p$, the warped coordinate $\mathbf{x}_{p}^{\mathcal{S}}$ can be calculated using a compression rate $\mathcal{C}<1$ as follows:
	\vspace{-0.1cm}
	\begin{equation}
		\footnotesize
		\begin{aligned}
			\mathcal{S}: \ \mathbf{x}_{p}^{\mathcal{S}} = \mathcal{C} \cdot \mathbf{x}_p.
		\end{aligned}
		\vspace{-0.1cm}
	\end{equation}
	\subsubsection{Unstructured Rich-textured Area}
	As shown in Fig. \ref{fig:quasi_pipeline}(c), if an area does not belong to the rich-textured area with low-frequency directions or the weak-textured area, it is identified as the unstructured rich-textured area. In this case, we apply the identity warping function $\mathcal{I}: \ \mathbb{R}^{3} \to \mathbb{R}^{3}$ to a given point $\mathbf{x}_p$ and the warped coordinate $\mathbf{x}_{p}^{\mathcal{I}}$ can be calculated as follows:
	\begin{equation}
		\footnotesize
		\begin{aligned}
			\mathcal{I}: \ \mathbf{x}_{p}^{\mathcal{I}} = \mathbf{x}_p.
		\end{aligned}
	\end{equation}
	\vspace{-1.0cm}
	\subsection{Hash Grid with Multiple Hash Functions}
	\label{subsec:quasi-h-hash}
	We employ uniform multiresolution hash grids\cite{muller2022instant} on the warped space to store the features. It works by arranging the neighboring grids of a specific sample point at $L$ resolution levels. At each level, $F$ dimensional features are assigned to the corners of the grids by looking up a hash table.
	However, it is possible for two different points in the original space to be mapped to the same position in the warped space. 
	Specifically,
	conflicts between $\mathbf{x}_{p}^{\mathcal{A}_{m}(v_i,t)}$ may arise within a specific voxel $v_i$ when multiple low-frequency directions exist (different $m$, same $v_i$). Besides, conflicts can also occur among $\mathbf{x}_{p}^{\mathcal{A}_{m}(v_i,t)}$, $\mathbf{x}_{p}^{\mathcal{S}}$ and $\mathbf{x}_{p}^{\mathcal{I}}$.
	However, as the compression rate $\mathcal{C}\!<\!1$ and the compression occurring around the voxel's center, conflicts are avoided between $\mathbf{x}_{p}^{\mathcal{A}_{m}(v_i,t)}$ in different voxels $v_i$ for a specific low-frequency direction (same $m$, different $v_i$) after affine warping. 
	Instead of building different grids for each warped space, we use a single hash table to index the grid vertices $\mathbf{\hat{v}}$ with various hash functions\cite{wang2023f2} conditioned on the types of space warping as follows:
	\vspace{-0.15cm}
	\begin{equation}
		\footnotesize
		\begin{aligned}
			{\rm Hash}_i(\hat{\mathbf{v}}) = \left(\bigoplus_{k=1}^{3}\hat{\mathbf{v}}_k\pi_{i,k} + \Delta_{i,k} \right)\mod T,
		\end{aligned}
		\vspace{-0.15cm}
	\end{equation}
	\begin{figure}[t!]
		\centering
		\includegraphics[width=1\linewidth]{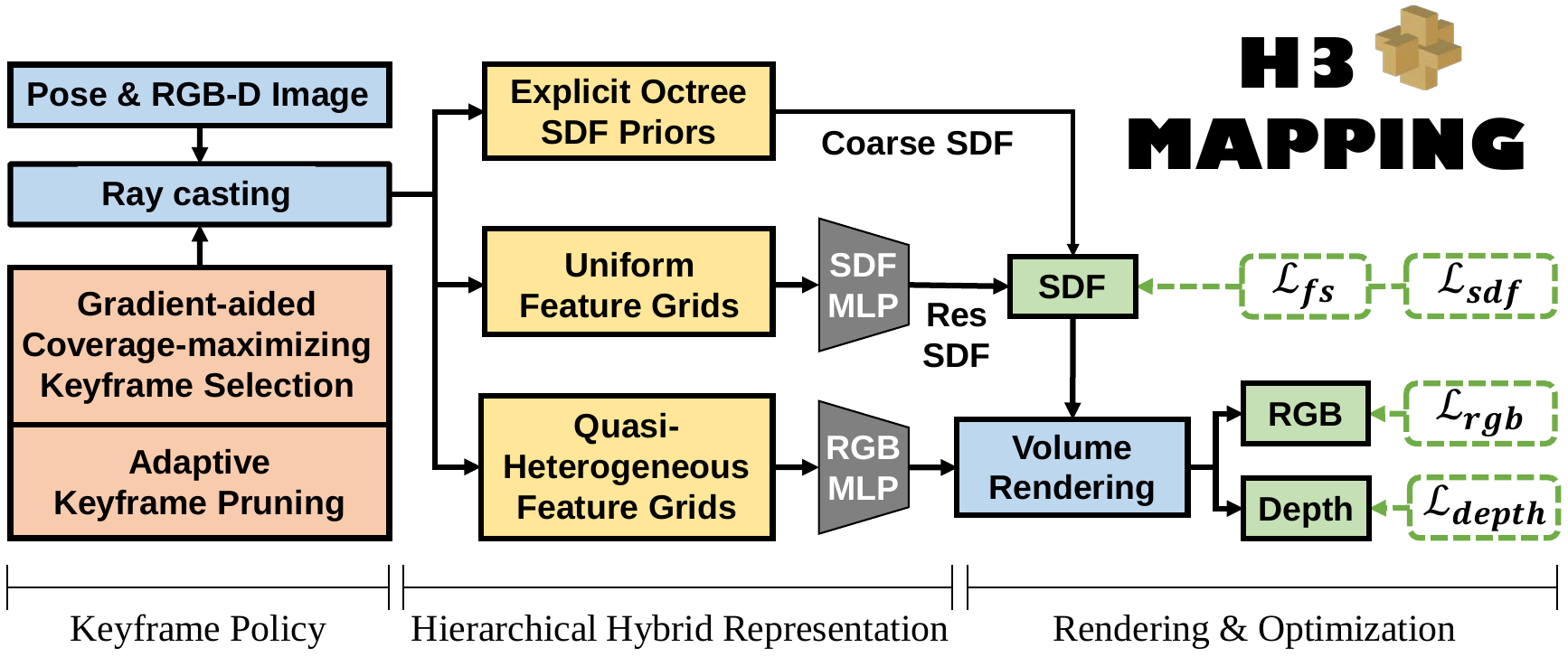}
		\caption{The pipeline of H3-Mapping.}
		\label{fig:mapping_pipeline}
		\vspace{-1.1cm}
	\end{figure}
	where $\bigoplus$ denotes the bitwise xor operation, and both $\left\{\pi_{i,k}\right\}$ and $\left\{\Delta_{i,k}\right\}$ are random large prime numbers, which are fixed for a specific warping space. $i = 1, 2, ..., M+2$ represents one scaling warping function $\mathcal{S}$, one identity warping function $\mathcal{I}$, and $M$ affine warping functions $\mathcal{A}_m$.
	Besides,  $k=1,2,3$ means the index of $x,y,z$ coordinate in the warped space, $T$ is the length of the hash table.
	The calculated hash value serves as an index for the hash table to retrieve features for the vertex $\mathbf{\hat{v}}$ and the features of the sample point are trilinear-interpolated from $8$ vertex features. 
	
	As depicted in Fig. \ref{fig:quasi_pipeline}(\uppercase\expandafter{\romannumeral4}), to ensure the same feature vector lengths for every sample point, we preallocate separate feature spaces for the $M\!+\!2$ hash functions to build the $(M\!+\!2)\! \times\! L \!\times\! F$ dimensional extended features. If a specific warping function exists, we insert the informative features obtained from the corresponding hash table. However, if a particular warping function does not exist, we insert zero-padded features as placeholders. 
	Due to the coexistence of multiple texture patterns at the same voxel (shown in Fig. \ref{fig:Quasi_Hetero_Gird}(b)), and the discontinuity of interpolated features at voxel edges (similar to NeuRBF\cite{chen2023neurbf}), the features obtained without space warping are always concatenated. This allows the network to select features accordingly to balance the fitting accuracy and interpolation smoothness.
	
	\vspace{-0.3cm}
	\section{H$_3$-Mapping}
	As outlined in Fig. \ref{fig:mapping_pipeline}, we propose a real-time and high-quality mapping method that utilizes the hierarchical hybrid representation (Sec. \ref{subsec:h2presentation}) with quasi-heterogeneous feature grids. Using the keyframe policy combining gradient-aided coverage-maximizing selection (Sec. \ref{subsec:key_slct}) and adaptive pruning (Sec. \ref{subsec:key_prune}), we use volume rendering like NeRF\cite{mildenhall2021nerf} to get depth and color of each sampled ray (Sec. \ref{subsec:sdf-volume-rendering}) and do the optimization using RGB-D inputs (Sec. \ref{subsec:optimization_process}). 
	\begin{table*}[]
		\centering
		\caption{Evaluation results of Replica\cite{straub2019replica}. The \textbf{\textcolor{darkred}{best}} and \textbf{second best} are separately marked in red bold and black bold.}
		\label{tab::replica_recon_render_metric}
		\resizebox{0.75\hsize}{!}{$
			\begin{tabular}{c||c||ccccccccc}
				\Xhline{2\arrayrulewidth}
				Metrics                                                                       & Method              & Room0          & Room1          & Room2          & Office0        & Office1        & Office2        & Office3        & Office4        & Avg.           \\ \hline
				\multirow{5}{*}{Depth L1${[}cm{]}$ $\downarrow$ }                                             
				& ESLAM                & 0.347  & 0.286 & 0.518 & 0.382 & 0.952 & 0.643 & 0.611 & 0.713 & 0.557          \\
				& Point-SLAM           & 0.357	& 0.202	& \textbf{0.447}	& \textbf{0.256}	& \textbf{\textcolor{darkred}{0.297}}	& \textbf{0.376}	& \textbf{0.466}	& \textbf{0.353}	& \textbf{0.344}   \\
				& SplaTAM-S                & 0.513  & 0.463 & 0.804 & 0.347 & 0.359 & 0.571 & 0.956 & 0.658 & 0.584          \\
				& H$_2$-Mapping        & \textbf{0.323}	& \textbf{0.194}	& 0.516	& 0.306	& 0.360	& 0.409	& 0.495	& 0.380 & 0.373   \\
				& \textbf{Ours}   & \textbf{\textcolor{darkred}{0.275}}	 & \textbf{\textcolor{darkred}{0.172}}	 & \textbf{\textcolor{darkred}{0.412}}	 & \textbf{\textcolor{darkred}{0.243}}	 & \textbf{0.323}	 & \textbf{\textcolor{darkred}{0.298}}	 & \textbf{\textcolor{darkred}{0.401}}	 & \textbf{\textcolor{darkred}{0.257}}	 & \textbf{\textcolor{darkred}{0.298}} \\ \hline
				\multirow{5}{*}{Accuracy${[}cm{]}$ $\downarrow$ }                                                 & ESLAM           & 1.210           & 0.977           & 1.052           & 0.943           & 1.200           & 1.213           & 1.351           & 1.352           & 1.159           \\
				& Point-SLAM           & \textbf{1.194}	& \textbf{0.940}	& \textbf{1.011}	& \textbf{\textcolor{darkred}{0.885}}	& \textbf{\textcolor{darkred}{0.700}}	& \textbf{1.079}	& \textbf{\textcolor{darkred}{1.287}}	& \textbf{1.226}	& \textbf{1.040} \\
				& SplaTAM-S                & 1.415  & 1.162 & 1.373 & 0.981 & 0.777 & 1.344 & 1.771 & 1.508 & 1.291          \\
				& H$_2$-Mapping      	& 1.197		& 0.943		& 1.026		& 0.909		& 0.725		& 1.081		& 1.300		& 1.227		& 1.051 \\
				& \textbf{Ours}           
				& \textbf{\textcolor{darkred}{1.189}} 	& \textbf{\textcolor{darkred}{0.938}}  	& \textbf{\textcolor{darkred}{1.006}} 	& \textbf{0.888}  	& \textbf{0.716} 	& \textbf{\textcolor{darkred}{1.063}}  	& \textbf{1.289} 	& \textbf{\textcolor{darkred}{1.202}}  	& \textbf{\textcolor{darkred}{1.036}}  \\ \hline
				\multirow{5}{*}{Completion${[}cm{]}$ $\downarrow$ }                                                & ESLAM           & 1.216           & 0.978           & 1.132           & 0.965           & 0.844           & 1.194           & 1.370           & 1.369           & 1.134           \\
				& Point-SLAM           & 1.239	& 0.968	& 1.140	& 0.958	& 0.787	& 1.149	& 1.364	& 1.303	& 1.114 \\
				& SplaTAM-S                & 1.351  & 1.144 & 1.389 & 1.015 & 0.831 & 1.313 & 1.686 & 1.798 & 1.316          \\
				& H$_2$-Mapping       & \textbf{1.206}	& \textbf{0.948}	& \textbf{1.104}	& \textbf{0.915}	& \textbf{0.760}	& \textbf{1.120}	& \textbf{1.322}	& \textbf{1.259}	& \textbf{1.079} \\
				& \textbf{Ours} & \textbf{\textcolor{darkred}{1.204}}	& \textbf{\textcolor{darkred}{0.945}}	& \textbf{\textcolor{darkred}{1.089}}	& \textbf{\textcolor{darkred}{0.902}}	& \textbf{\textcolor{darkred}{0.754}}	& \textbf{\textcolor{darkred}{1.091}}	& \textbf{\textcolor{darkred}{1.306}}	& \textbf{\textcolor{darkred}{1.246}}	& \textbf{\textcolor{darkred}{1.067}}  \\ \hline
				\multirow{5}{*}{\begin{tabular}[c]{@{}c@{}}SSIM $\uparrow$ \end{tabular}} & ESLAM           & 0.842           & 0.876           & 0.901           & 0.879           & 0.932           & 0.890           & 0.897           & 0.924           & 0.893           \\
				& Point-SLAM     & 0.855          & 0.889          & 0.918          & 0.930          & 0.932          & 0.891          & 0.889          & 0.910          & 0.902          \\
				& SplaTAM-S                & \textbf{0.902}  & \textbf{0.937} & \textbf{0.946} & \textbf{\textcolor{darkred}{0.969}} & 0.955 & \textbf{0.931} & 0.914 & 0.933 & \textbf{0.936}          \\
				& H$_2$-Mapping         & 0.893         & 0.923         & 0.929         & \textbf{0.962}         & \textbf{0.957}         & 0.928         & \textbf{0.924}         & \textbf{0.940}         & 0.932 \\
				& \textbf{Ours}         & \textbf{\textcolor{darkred}{0.921}}         & \textbf{\textcolor{darkred}{0.939}}         & \textbf{\textcolor{darkred}{0.948}}         & \textbf{\textcolor{darkred}{0.969}}         & \textbf{\textcolor{darkred}{0.966}}         & \textbf{\textcolor{darkred}{0.936}}         & \textbf{\textcolor{darkred}{0.933}}         & \textbf{\textcolor{darkred}{0.950}}         & \textbf{\textcolor{darkred}{0.945}} \\ \hline
				\multirow{5}{*}{\begin{tabular}[c]{@{}c@{}}PSNR$[db]$ $\uparrow$\end{tabular}} 
				& ESLAM          & 29.49          & 30.65          & 31.50          & 30.35          & 36.40          & 30.19          & 31.12          & 32.72          & 31.55          \\
				& Point-SLAM     & 29.58          & 30.89          & 32.89          & 35.06          & 35.90          & 31.02          & 30.97          & 31.36          & 32.21          \\
				& SplaTAM-S                & 31.30  & 33.45 & \textbf{34.33} & \textbf{39.23} & \textbf{39.04} & 31.97 & 30.74 & 33.23 & 34.16          \\
				& H$_2$-Mapping          & \textbf{31.76}          & \textbf{33.61}          & 32.89          & 38.67          & 38.92          & \textbf{32.68}          & \textbf{33.13}          & \textbf{34.34}          & \textbf{34.49}          \\
				& \textbf{Ours}         & \textbf{\textcolor{darkred}{33.16}}         & \textbf{\textcolor{darkred}{34.99}}         & \textbf{\textcolor{darkred}{35.24}}         & \textbf{\textcolor{darkred}{39.85}}         & \textbf{\textcolor{darkred}{40.12}}         & \textbf{\textcolor{darkred}{33.89}}         & \textbf{\textcolor{darkred}{34.10}}         & \textbf{\textcolor{darkred}{35.99}}         & \textbf{\textcolor{darkred}{35.92}} \\ \hline
			\end{tabular}$}
		\vspace{-0.9cm}
	\end{table*}
	\vspace{-0.3cm}
	\subsection{Hierarchical Hybrid Representation}
	\label{subsec:h2presentation}
	To accurately reconstruct the scene geometry, we use the efficient hybrid representation introduced in our previous work\cite{jiang2023h2}. Specifically, we store the optimizable SDF in each leaf node's vertex of the octree to represent the coarse geometry, which can be initialized by projecting it onto the input depth image. The coarse SDF $s^c$ of any sample point is obtained from its surrounding eight vertices through tri-linear interpolation.
	To capture geometry details, we use the uniform multiresolution feature grids\cite{muller2022instant} to solely address the residual geometry. It is much simpler
	to learn than the complete geometry, thereby improving the
	geometry accuracy and convergence rate. By inputting the features into a small MLP, the SDF prediction $s \in \mathbb{R}^1$ is got as follows:
	\vspace{-0.1cm}
	\begin{equation}
		\footnotesize
		\begin{aligned}
			s = s^{c} + \mathcal{M}_s(\phi ^{s};\theta^{w}_{s})
		\end{aligned},
		\vspace{-0.2cm}
	\end{equation}
	where $\phi ^{s}$ are $L\! \times\! F$ dimensional features obtained from the uniform multiresolution feature grids. $\mathcal{M}_s$, parameterized by $\theta^{w}_{s}$, is the MLP to output the residual SDF prediction.
	
	For texture modeling, we use quasi-heterogeneous feature grids and employ a tiny MLP for color prediction as follows:
	\vspace{-0.3cm}
	\begin{equation}
		\footnotesize
		\begin{aligned}
			\mathbf{c} = \mathcal{M}_c(\phi ^{c};\theta^{w}_{c}),
		\end{aligned}
		\vspace{-0.2cm}
	\end{equation}
	where  $\phi ^{c}$  is the $(M+2) \!\times\! L \!\times\! F$ dimensional extended features from the quasi-heterogeneous feature grids. $\mathcal{M}_c$, parameterized by $\theta^{w}_{c}$, is the MLP to output the color prediction $\mathbf{c} \in \mathbb{R}^3$.
	\begin{figure}[t!]
		\centering
		\includegraphics[width=1\linewidth]{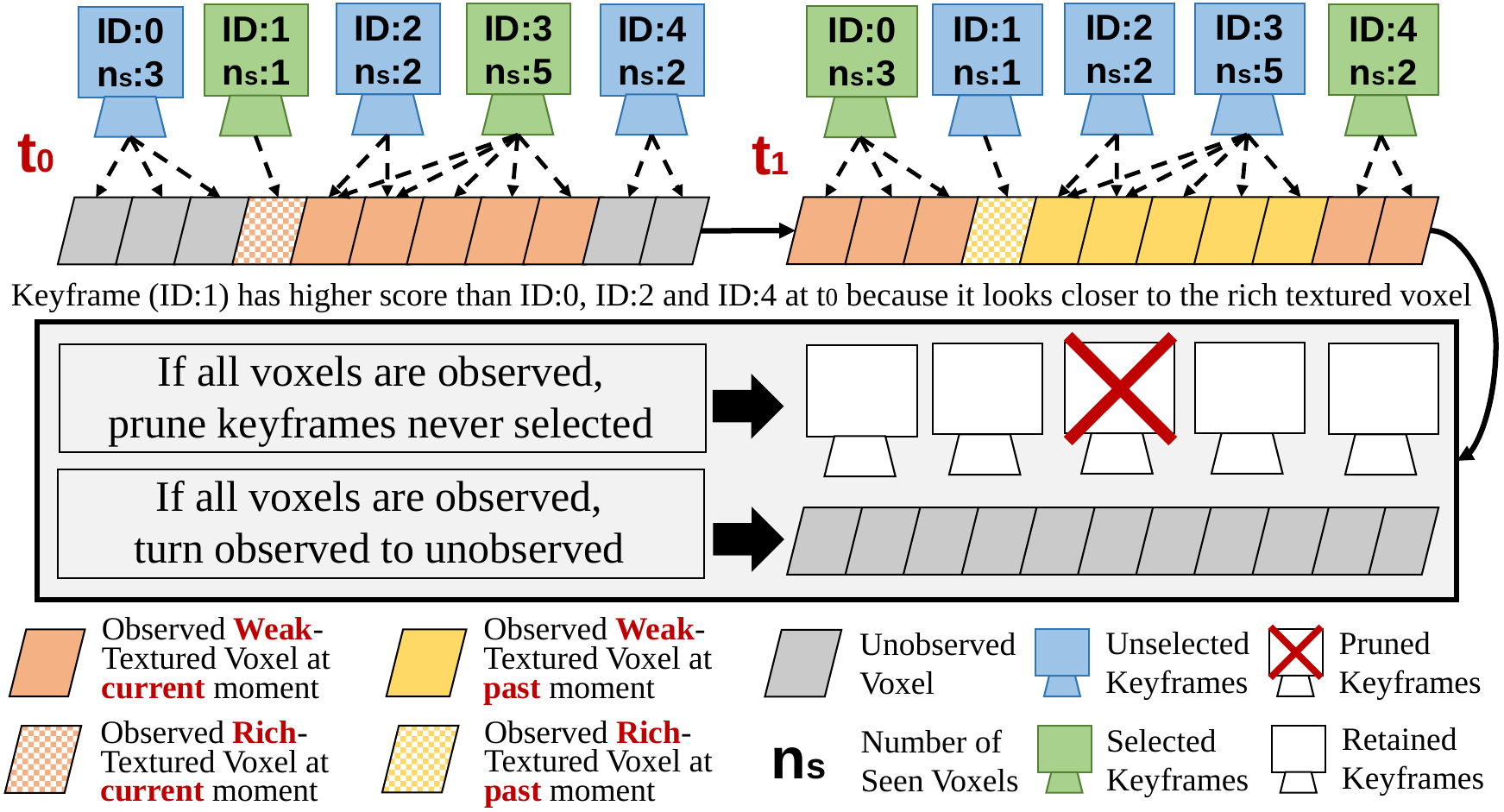}
		\caption{In each training iteration, we select $K$ keyframes that cover the voxels with the top keyframe scores and remove any keyframes that have not been utilized when all voxels have been observed. Assuming $K\!=\!2$ for simplicity.}
		\label{fig:kf_policy}
		\vspace{-1.1cm}
	\end{figure}
	\vspace{-0.7cm}
	\subsection{Keyframe Policy}
	\subsubsection{Gradient-aided Coverage-maximizing Strategy for Keyframe Selection}
	\label{subsec:key_slct}
	For a new input RGB-D frame, we insert the frame into the keyframe set if the overlapped voxel ratio with the last inserted frame is lower than $0.85$, or 10 frames have passed since the last keyframe insertion. 
	During each training iteration, we utilize the color gradient to select keyframes that have a closer focus on rich-textured areas and a broader scope for weak-textured regions. Specifically, we define the score of each voxel $v_i$ as follows:
	\vspace{-0.1cm}
	\begin{equation}
		\footnotesize
		\begin{aligned}
			score(v_i) = max(cnt^{2}(v_i, f_j)  g(v_i, f_j), 1),
		\end{aligned}
		\vspace{-0.1cm}
	\end{equation}
	where all the symbols are denoted in \ref{subsubsec:weak}. 
	It implies that looking closer (increasing $cnt^{2}(v_i, f_j)$) to the rich-textured areas (larger $g(v_i, f_j)$) will have a greater impact than weak-textured areas and the corresponding frames are more likely to be selected. As shown in Fig. \ref{fig:kf_policy}, we begin by selecting $K$ keyframes that cover the voxels with the highest sum of scores. After labeling the covered voxels as observed we use the same selection strategy but only consider the remaining unobserved voxels in the next time step. If all voxels have been labeled as observed, the process is repeated by resetting the voxels to be labeled as unobserved.
	\subsubsection{Adaptive Keyframe Pruning}
	\label{subsec:key_prune}
	As shown in Fig. \ref{fig:kf_policy}, when all voxels have been labeled as observed, we remove the frames that were not selected before. This approach allows us to retain as much historical information as possible while maintaining storage efficiency.
	\vspace{-0.3cm}
	\subsection{SDF-based Volume rendering}
	\label{subsec:sdf-volume-rendering}
	Like Vox-Fusion\cite{yang2022vox}, we only sample points along the ray that intersects with any leaf node voxel of octree. And then get rendered color $\mathbf{C}$ and depth $D$ for each ray as follows:
	\vspace{-0.1cm}
	\begin{equation}
		\footnotesize
		\resizebox{0.85\hsize}{!}{$
			\begin{aligned}
				\alpha_{j} = \sigma(\frac{s_j}{\beta}) \cdot \sigma(-\frac{s_j}{\beta}),  \ 
				\mathbf{C} = \frac{1}{\sum^{N-1}_{j=0}\alpha_j}\sum^{N-1}_{j=0}\alpha_j\cdot\mathbf{c}_j, \  
				D = \frac{1}{\sum^{N-1}_{j=0}\alpha_j}\sum^{N-1}_{j=0}\alpha_j\cdot d_j, 
			\end{aligned}
			$}
		\vspace{-0.1cm}
	\end{equation}
	where $\sigma(\cdot)$ is a sigmoid function, $s_{j}$ and $\mathbf{c}_j$ are the predicted SDF and color, $N$ is the number of samples along the ray, $\beta$ is a truncation distance and $d_{j}$ is the sample's depth.
	\vspace{-0.3cm}
	\subsection{Optimization Process}
	\label{subsec:optimization_process}
	We apply RGB Loss ($\mathcal{L}_{rgb}$), Depth Loss ($\mathcal{L}_{d}$), Free Space Loss ($\mathcal{L}_{fs}$) and SDF Loss ($\mathcal{L}_{sdf}$) on a batch of rays $R$ for optimization. The final loss function is defined as:
	\vspace{-0.1cm}
	\begin{equation}
		\footnotesize
		\begin{aligned}
			\mathcal{L} = \alpha_{sdf}\mathcal{L}_{sdf} + \alpha_{fs}\mathcal{L}_{fs} +\alpha_{d}\mathcal{L}_{d} +\alpha_{rgb}\mathcal{L}_{rgb}, 
		\end{aligned}
		\vspace{-0.1cm}
	\end{equation}
	where $\alpha_{sdf}$, $\alpha_{fs}$ ,$\alpha_{d}$, and $\alpha_{rgb}$ are the weighting coefficients.
	\subsubsection{Rendering Loss}
	\begin{equation}
		\footnotesize
		\resizebox{0.85\hsize}{!}{$
			\begin{aligned}
				\mathcal{L}_{\!d\!} = \frac{1}{\vert R\vert}\!\sum_{\!r\in R}\lVert D_r\! -\! D^{gt}_r\rVert ,  \ 
				\mathcal{L}_{\!rgb\!} = \frac{1}{\vert R\vert}\!\sum_{\!r\in R}\lVert \mathbf{C}_r\! -\! \mathbf{C}^{gt}_r\rVert + \text{S3IM}(\mathbf{C}_r,\mathbf{C}^{gt}),
			\end{aligned}
			$}
		\vspace{-0.1cm}
	\end{equation}
	where ($D_{r}$, $D^{gt}_r$) and ($\textbf{C}_{r}$, $\mathbf{C}^{gt}_r$) are rendered and input depth and color. S3IM\cite{xie2023s3im} is the stochastic structural similarity between two groups of pixels.
	\subsubsection{Surface-aware TSDF Loss}
	\begin{equation}
		\footnotesize
		\resizebox{0.85\hsize}{!}{$
			\begin{aligned}
				\mathcal{L}_{\!fs\!}\! =\! \frac{1}{\vert R\vert }\!\sum_{\!r\in R}\! \frac{1}{P_r^{fs}}\!\sum_{\mathbf{x}_{p} \in P_r^{fs}}\!(s_{\mathbf{x}_{p}}\! -\! tr\!)^2 ,  \ 
				\mathcal{L}_{\!sdf\!}\! =\! \frac{1}{\vert R\vert }\!\sum_{\!r\in R}\! \frac{1}{P_r^{tr}}\!\sum_{\mathbf{x}_{p} \in P_r^{tr}}\!(s_{\mathbf{x}_{p}}\! -\! s_{\mathbf{x}_{p}}^{gt}\!)^2 ,
			\end{aligned}
			$}
		\vspace{-0.1cm}
	\end{equation}
	where $s_{\mathbf{x}_{p}}$ is the predicted SDF and $s_{\mathbf{x}_{p}}^{gt}$ is the difference between the distance to point $\mathbf{x}_{p}$ on the ray $r$ and the depth measurement of that ray. $P_r^{fs}$ is a set of points on the ray $r$ that lies between the camera and the truncation region of the surface based on $D^{gt}_r$. $P_r^{tr}$ is a set of points within the truncation area. 
	For the objects smaller than the truncation region, the negative $s_{\mathbf{x}_{p}}^{gt}$ used to supervise the inside object region can be inaccurate. Therefore, to ensure that $\mathcal{L}_{\!sdf\!}$ is constrained to a single surface, we eliminate negative $s_{\mathbf{x}_{p}}^{gt}$ supervision for points where a sign change in the SDF occurs for the second time along the ray.
	\begin{figure*}[t!]
		\centering
		\includegraphics[width=1\textwidth]{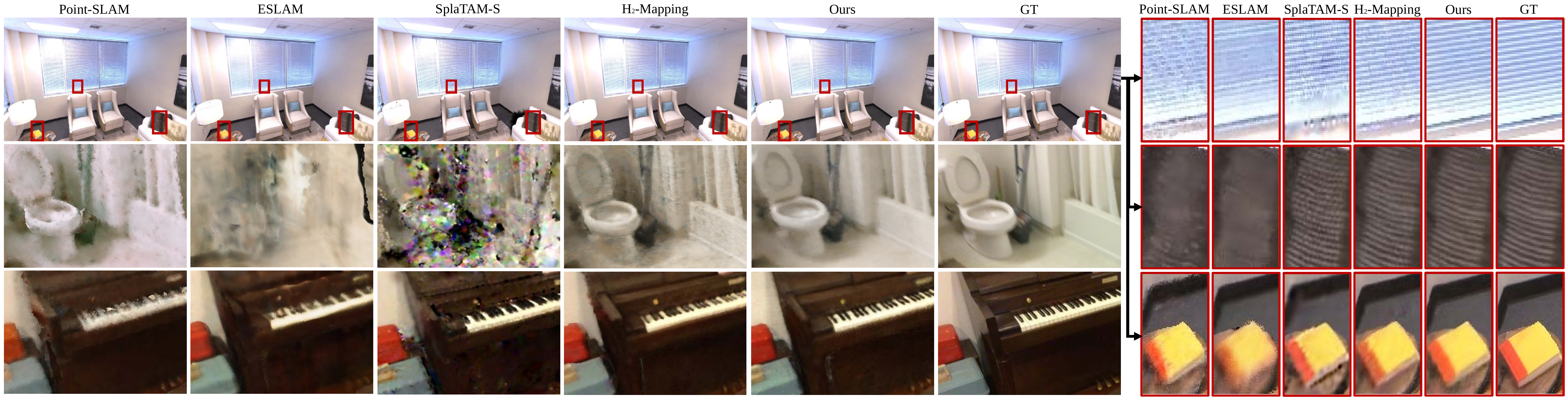}
		\caption{Rendering results of Replica\cite{straub2019replica} and ScanNet\cite{dai2017scannet}. The magnification of the red box areas is shown on the right.}
		\label{fig:exp_render}
		\vspace{-0.7cm}
	\end{figure*}
	\vspace{-0.3cm}
	\section{Experiment}
	\subsection{Implementation Details} 
	In our method, the parameters of the hierarchical hybrid representation align with H$_2$-Mapping \cite{jiang2023h2}, except for a scale difference is set to $3$ for texture in the hash grids\cite{muller2022instant}. This means that we do not increase the number of multiresolution hash grids than H$_2$-Mapping \cite{jiang2023h2}.
	Besides, we perform $2$ iterations per frame, selecting $8192$ pixels per iteration for training. The texture pattern is determined every $10$ frames. The maximum low-frequency direction of $M$ is set to $2$, and the compression rate $\mathcal{C}$ is set to $0.1$. We use threshold values of $tr_{plane} = 0.01m$, $tr_{near} = 0.95$, and $tr_{weak} = 0.2$. These configurations remain consistent across all datasets. 
	\vspace{-0.5cm}
	\subsection{Mapping and Rendering Evaluation}
	\subsubsection{Baselines} 
	We select three advanced NeRF-based dense RGB-D mapping methods currently open-source, ESLAM\cite{johari2022eslam}, Point-SLAM\cite{sandstrom2023point}, and H$_2$-Mapping\cite{jiang2023h2}, along with the gaussian-splatting-based method SplaTAM\cite{keetha2023splatam} for comparison. However, since we solely focus on incremental mapping, we remove the tracking component by providing the ground truth pose and use every frame for training. Since the default mapping iterations of Point-SLAM\cite{sandstrom2023point} are set too high, resulting in extremely slow performance, we reduce its maximum iterations to be the same as ESLAM\cite{johari2022eslam} for fair comparison.
	Besides, we use the SplaTAM-S\cite{keetha2023splatam} configuration due to its faster performance. To generate the mesh of SplaTAM-S\cite{keetha2023splatam} for geometric evaluation, we use the same method employed in Point-SLAM\cite{sandstrom2023point} to generate the mesh from rendered depth images.
	All other aspects remain unchanged.
	\subsubsection{Metrics} 
	\label{subsubsec:metrics}
	Before evaluation, we remove faces from a mesh that are not inside any camera frustum or are occluded in all input frames like ESLAM\cite{johari2022eslam}.
	To evaluate the geometry, we use the Depth L1 Error$[cm]$, Accuracy$[cm]$, and Completion$[cm]$ of the reconstructed mesh with the same resolution. 
	Besides, we use structural similarity index measure (SSIM) and peak signal-to-noise ratio (PSNR) to evaluate texture on rendered images. 
	\begin{table}[t!]
		\centering
		\caption{Runtime Analysis.}
		\label{tab:run_time}
		\begin{tabular}{c||cccc}
			\hline
			\multirow{3}{*}{Method} & \multicolumn{4}{c}{Average Frame Processing Time (s)}                          \\ \cline{2-5} 
			& \multicolumn{2}{c|}{Replica: Average} & \multicolumn{2}{c}{ScanNet: Scene0207} \\
			& RTX 4090  & \multicolumn{1}{c|}{AGX Orin} & RTX 4090             & AGX Orin            \\ \hline
			ESLAM                   & 0.178 & \multicolumn{1}{c|}{1.093}    & 0.488            & 2.779               \\
			Point-SLAM              & 0.692 & \multicolumn{1}{c|}{2.848}    & 1.149            & 8.944               \\
			SplaTAM-S                         & \multicolumn{1}{c}{0.366} & \multicolumn{1}{c|}{2.651} & \multicolumn{1}{c}{0.679} & \multicolumn{1}{c}{4.299} \\
			H$_2$-Mapping & \multicolumn{1}{c}{0.059} & \multicolumn{1}{c|}{0.362} & \multicolumn{1}{c}{0.066} & \multicolumn{1}{c}{0.501} \\
			\textbf{Ours}                    & \textbf{0.053} & \multicolumn{1}{c|}{\textbf{0.308}}    & \textbf{0.059}            & \textbf{0.426}               \\ \hline
		\end{tabular}
		\resizebox{0.95\linewidth}{!}{$
			\begin{tabular}{ccccc}
				\\
				\hline
				\multicolumn{5}{c}{Replica: Average Processing Time (ms per frame(f)) on RTX 4090} \\ \hline
				\begin{tabular}[c]{@{}c@{}}Data Reading \\ \& Prepocessing\end{tabular} &
				\begin{tabular}[c]{@{}c@{}}Octree \\ SDF priors\end{tabular} &
				\begin{tabular}[c]{@{}c@{}}Texture Pattern\\ Determination\end{tabular} &
				\begin{tabular}[c]{@{}c@{}}Keyframe\\ Policy\end{tabular} &
				\begin{tabular}[c]{@{}c@{}}Map\\ Training\end{tabular} \\ \hline
				16.65/f &
				8.305/f &
				10.02/(10f) &
				1.943/f &
				27.10/f \\ \hline
			\end{tabular}$}
		\vspace{-1.4cm}
	\end{table}
	\subsubsection{Evaluation on Replica\cite{straub2019replica}}
	In \autoref{tab::replica_recon_render_metric}, we present a quantitative comparison of the geometry accuracy and rendering fidelity between our method and the baselines. The results show that our approach outperforms the baselines for both averaged 2D and 3D metrics. It is noted that Point-SLAM\cite{sandstrom2023point} uses ground-truth depth for depth rendering, resulting in higher geometric accuracy in some scenes. Besides, we provide a qualitative analysis of Room0 in the first row of Fig. \ref{fig:exp_render}. It demonstrates that our method can generate renderings with more realistic details on blinds, pillows, and books.
	\subsubsection{Evaluation on ScanNet\cite{dai2017scannet}}
	Due to inaccurate ground truth meshes and blurred input images in the ScanNet dataset\cite{dai2017scannet}, we only provide qualitative analysis on Scene0050 and Scene0207 as shown in Fig. \ref{fig:exp_render}. Compared to the baselines, our method exhibits sharper object outlines, reduced noise, and higher-fidelity textures, like piano keys and toilets. More qualitative comparisons are available in Appendix.A\cite{suppmat_h3}.
	\subsubsection{Runtime Analysis}
	In \autoref{tab:run_time}, we evaluate the average frame processing time $[s]$ on an RTX 4090 and a Jetson AGX Orin across eight scenes in Replica\cite{straub2019replica} and Scene0207 in ScanNet\cite{dai2017scannet}. The results show our method is faster than the baselines. Additionally, we evaluate the average time required for each component in our method on RTX 4090.
	\vspace{-0.4cm}
	\subsection{Ablation Study}
	\subsubsection{Quasi-heterogeneous Feature Grids}
\begin{figure}[t!]
		\centering
		\includegraphics[width=1\linewidth]{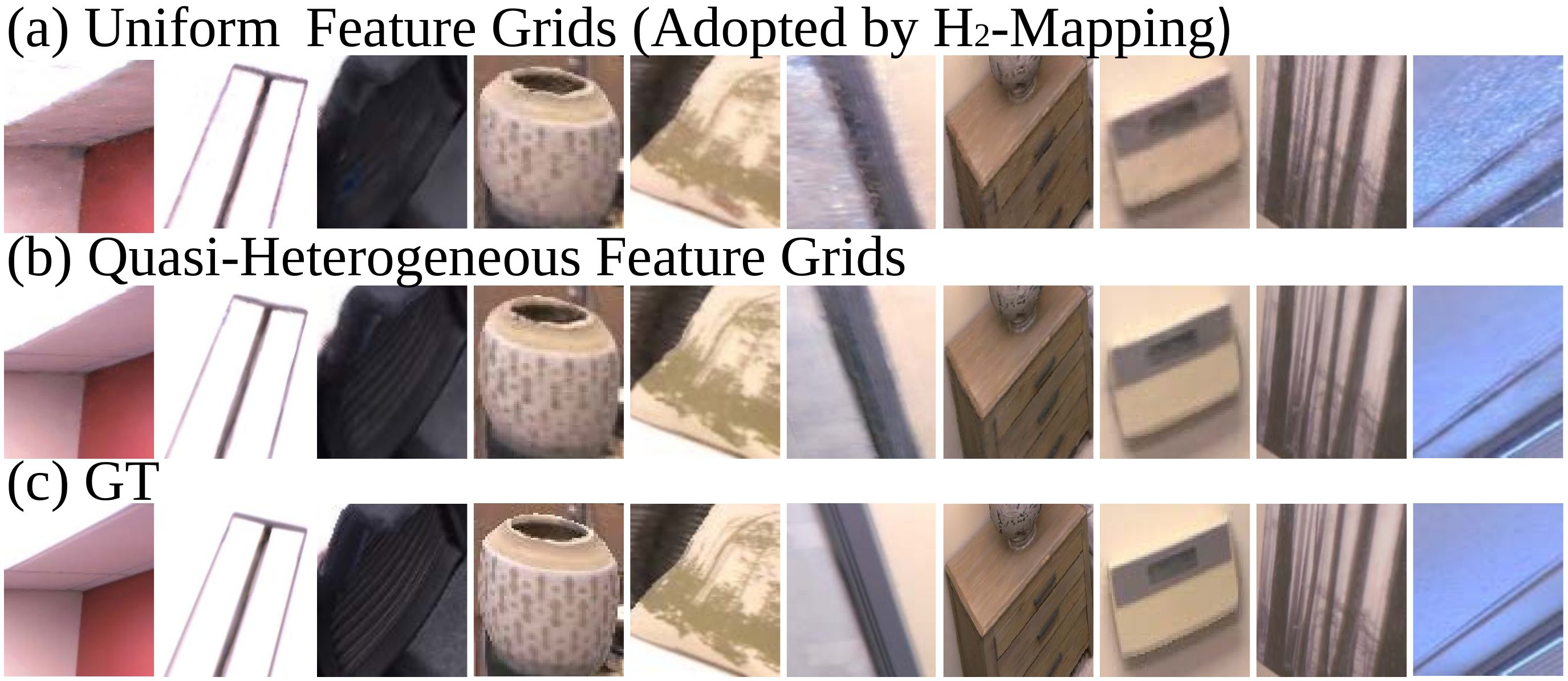}
		\caption{Comparison of rendering using quasi-heterogeneous feature grids versus uniform feature grids. Complete images are in Appendix.C\cite{suppmat_h3}}
		\label{fig:quasi_ablation}
		\vspace{-1.1cm}
	\end{figure}

	The use of quasi-heterogeneous feature grids adapts to varying levels of texture complexity in the scene, resulting in more efficient texture optimization with fewer samples and reduced training time. As shown in Fig.\ref{fig:iter_curve}(c), texture optimization in our method achieves higher fidelity with fewer iterations compared to methods that use uniform feature grids. 
	Additionally, as shown in \autoref{tab:ablation_quais} (with detailed results for each scene available in Appendix.B\cite{suppmat_h3}), incorporating space warping for rich-textured areas with low-frequency directions and weak-textured areas can both lead to more accurate texture modeling. Since the scene geometry continues to be represented by uniform feature grids, geometric accuracy remains unchanged. Fig.\ref{fig:quasi_ablation} also shows that quasi-heterogeneous feature grids provide more details in rich-textured areas and less noise in weak-textured areas.
	\begin{table}[t!]
		\centering
		\caption{Analysis of quasi-heterogeneous feature grids. The results are averaged over Replica dataset\cite{straub2019replica}. “w.R” and “w.W” are short for “Using spacing warping for rich-textured area with low-frequency directions” and “Using spacing warping for weak-textured area” respectively.}
		\label{tab:ablation_quais}
		%	\resizebox{\linewidth}{!}{
			\begin{tabular}{c||c||c|c|c}
				\Xhline{2\arrayrulewidth}
				w.R  & w.W    
				& PSNR $\uparrow [db]$   & Acc. $[cm] \ \downarrow$ & Comp. $[cm] \ \downarrow$ \\ \hline
				$\times$   & $\times$  & 34.94  & 1.038 & 1.068          \\
				\checkmark & $\times$    & 35.65  & 1.037 & 1.068         \\
				$\times$ & \checkmark    & 35.39  & 1.037 & \textbf{1.067}         \\
				\checkmark & \checkmark  & \textbf{35.92} & \textbf{1.036} & \textbf{1.067} 
				\\ \Xhline{2\arrayrulewidth}
			\end{tabular}
			%	}
		\vspace{-0.3cm}
	\end{table}
	\subsubsection{Keyframe Policy}
\begin{figure}[t!]
		\centering
		\includegraphics[width=1\linewidth]{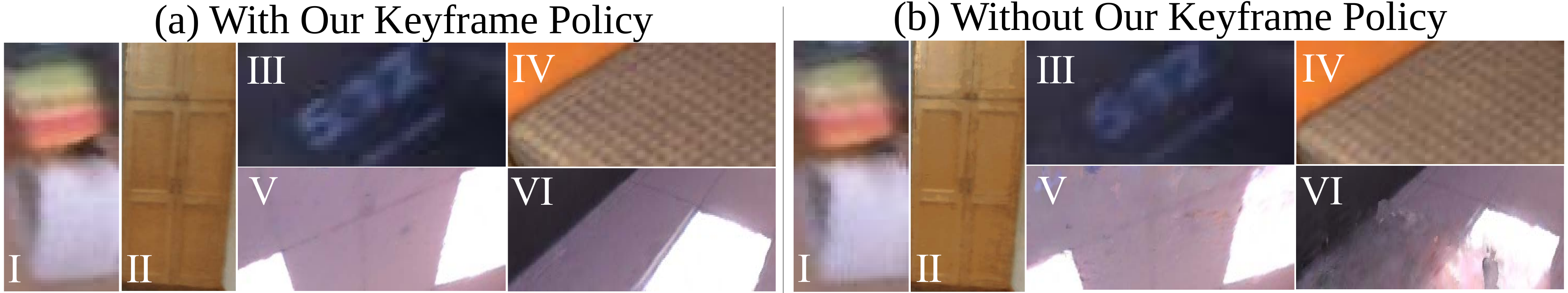}
		\caption{Rendering results with and without our keyframe policy for rich-textured regions and marginal areas. Complete images are in Appendix.C\cite{suppmat_h3}}
		\label{fig:insert_clean_kf}
		\vspace{-0.4cm}
	\end{figure}

	Compared to the one used in H$_2$-Mapping\cite{jiang2023h2}, the proposed gradient-aided coverage-maximizing strategy for keyframe selection additionally takes the texture complexity into account, resulting in higher PSNR in \autoref{tab:ablation_kf_sdf} and higher details as shown in the Fig. \ref{fig:insert_clean_kf}(I,II,III,IV).
	Since this strategy does not consider the geometry, it offers nearly no improvement in geometric accuracy. 
	Besides, the use of adaptive pruning allows for a decrease in the maximum insertion interval from $100$ to $10$, which leads to improved geometry and texture without significantly affecting the average number of keyframes (originally 131 frames, now 128 frames) in the Replica dataset\cite{straub2019replica}, as illustrated in \autoref{tab:ablation_kf_sdf}. 
	As shown in the Fig. \ref{fig:insert_clean_kf}(V,VI), the marginal areas can be better reconstructed.
	\subsubsection{Surface-aware TSDF Loss}
\begin{figure}[t!]
		\centering
		\includegraphics[width=1\linewidth]{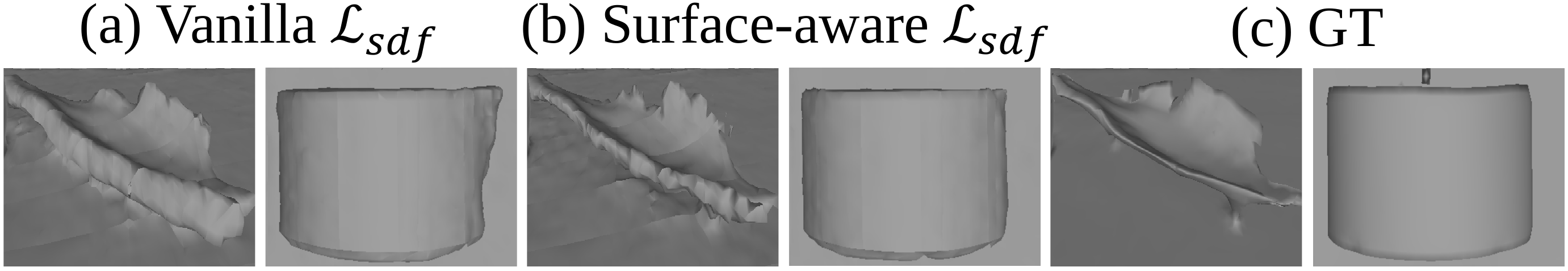}
		\caption{Reconstructed mesh with surface-aware and vanilla TSDF loss. Complete images are in Appendix.C\cite{suppmat_h3}}
		\label{fig:surface_aware_sdf}
		\vspace{-0.4cm}
	\end{figure}

	As shown in Table \ref{tab:ablation_kf_sdf} and Fig. \ref{fig:surface_aware_sdf}, by avoiding the negative SDF supervision behind the second surface, we observe improved mesh reconstruction with fewer artifacts and sharper outliers. In turn, better geometry enhances the texture modeling.
	\vspace{-0.5cm}
	\subsection{Real-world SLAM Demonstration}
\begin{figure}[t!]
		\centering
		\includegraphics[width=1\linewidth]{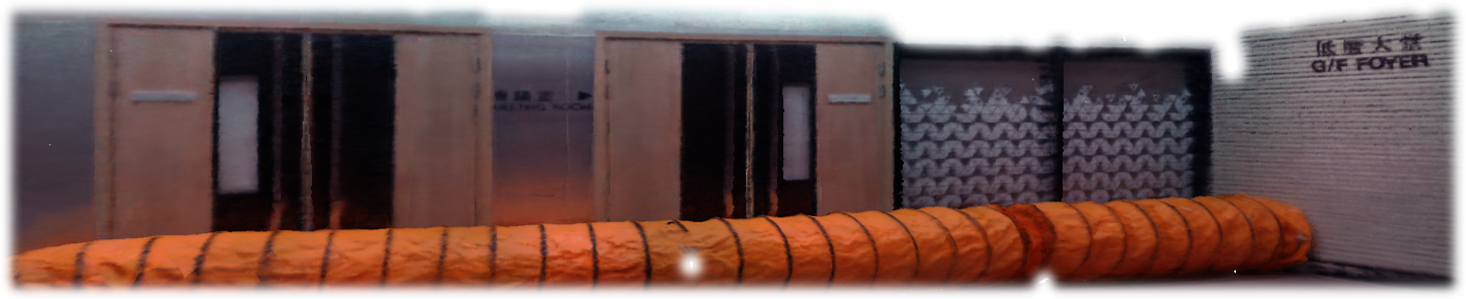}
		\caption{The textured mesh reconstructed in our real-world SLAM demonstration using a handheld device.}
		\label{fig:handheld}
		\vspace{-2.0cm}
	\end{figure}

	We achieve a SLAM system on a handheld device powered by AGX Orin by combining our mapping method with a tracking module. The reconstruction results are shown in the Fig. \ref{fig:handheld}, and more details can be found in the attached video.
	\begin{table}[t!]
		\centering
		\caption{Analysis of keyframe policy and surface-aware TSDF loss. Results are averaged over Replica dataset\cite{straub2019replica}.}
		\label{tab:ablation_kf_sdf}
		\resizebox{\linewidth}{!}{
			\begin{tabular}{c||c||c||ccc}
				\Xhline{2\arrayrulewidth}
				\shortstack{\\Surface-aware \\TSDF Loss}  &
				\shortstack{\\Keyframe \\ Selection}  & \shortstack{\\Adaptive \\ Pruning}    
				& \shortstack{\\PSNR$\uparrow$\\$[db]$}     & \shortstack{\\Acc.$\downarrow$\\$[cm]$}   & \shortstack{\\Comp.$\downarrow$\\$ [cm]$}   \\ \hline
				\checkmark   &\checkmark     & $\times$   & 35.75  & 1.039   & 1.072       \\
				\checkmark   &$\times$ & \checkmark  & 35.68  & 1.037   & 1.068      \\
				$\times$   &\checkmark   &\checkmark   & 35.55 & 1.051 & 1.075 \\
				\checkmark   &\checkmark  & \checkmark   & \textbf{35.92} & \textbf{1.036} & \textbf{1.067} \\ \Xhline{2\arrayrulewidth}
			\end{tabular}
		}
		\vspace{-0.4cm}
	\end{table}
	\section{Conclusion}
	We propose H3-Mapping, a novel NeRF-based dense mapping system that leverages a hierarchical hybrid representation with quasi-heterogeneous feature grids, resulting in real-time performance and high-quality reconstruction. By preserving the efficient uniform feature grids, we apply different space warping based on the analysis of varied texture patterns to reduce the redundant feature grid allocation. As a result, texture can be optimized effectively with limited sampling and training time. Additionally, we introduce a gradient-aided coverage-maximizing strategy for keyframe selection to enable optimized keyframes to achieve a closer focus on rich-textured areas and a broader look at weak-textured regions. This strategy mitigates the forgetting issue and enhances the mapping quality in both geometry and texture. 
	Baseline comparisons show that our method outperforms existing approaches in terms of mapping quality and time consumption. Ablation studies further confirm the effectiveness of each proposed module. However, our method cannot handle dynamic objects and pose drifting currently.
	\vspace{-0.5cm}
	
	\newlength{\bibitemsep}\setlength{\bibitemsep}{0\baselineskip}
	\newlength{\bibparskip}\setlength{\bibparskip}{0pt}
	\let\oldthebibliography\thebibliography
	\renewcommand\thebibliography[1]{%
		\oldthebibliography{#1}%
		\setlength{\parskip}{\bibitemsep}%
		\setlength{\itemsep}{\bibparskip}%
	}

	\bibliography{RAL2023}

\end{document}